\documentclass[journal,10pt]{elsarticle}

\usepackage{algorithmic}
\usepackage{array}

\usepackage{framed,multirow,booktabs}
\usepackage{graphicx,epsfig}
\usepackage{amssymb,amsmath,bm}
\usepackage{textcomp}
\usepackage{amsthm}
\usepackage{lineno,hyperref}
\usepackage{xcolor}
\usepackage{cleveref}
\usepackage[section]{placeins}

\usepackage{CJK}
\usepackage{makecell}
\usepackage{geometry}
\usepackage{setspace} 
\journal{Journal of Pattern Recognition}

\begin{document}
	\begin{frontmatter}
		%
		\title{Count, Decode and Fetch: A New Approach to Handwritten Chinese Character Error Correction}

		\author{Pengfei Hu, Jiefeng Ma, Zhenrong Zhang, Jun Du and Jianshu Zhang}

		\begin{abstract}
			Recently, handwritten Chinese character error correction has been greatly improved by employing encoder-decoder methods to decompose a Chinese character into an ideographic description sequence (IDS). 
			However, existing methods implicitly capture and encode linguistic information inherent in IDS sequences, leading to a tendency to generate IDS sequences that match seen characters. This poses a challenge when dealing with an unseen misspelled character, as the decoder may generate an IDS sequence that matches a seen character instead.
			Therefore, we introduce Count, Decode and Fetch (CDF), a novel approach that exhibits better generalization towards unseen misspelled characters. CDF is mainly composed of three parts: the counter, the decoder, and the fetcher. In the first stage, the counter predicts the number of each radical class without the symbol-level position annotations. In the second stage, the decoder employs the counting information and generates the IDS sequence step by step. Moreover, by updating the counting information at each time step, the decoder becomes aware of the existence of each radical. With the decomposed IDS sequence, we can determine whether the given character is misspelled. If it is misspelled, the fetcher under the transductive transfer learning strategy predicts the ideal character that the user originally intended to write. 
			We integrate our method into existing encoder-decoder models and significantly enhance their performance.

		\end{abstract}
		
		\begin{keyword}
			Handwritten Chinese character error correction \sep Radical \sep Zero-shot learning \sep Transfer learning
		\end{keyword}

	\end{frontmatter}
	{\bf Correspondence:}
	Dr. Jun Du, National Engineering Research Center for Speech and Language Information Processing (NERC-SLIP), University of Science and Technology of China, No. 96, JinZhai Road, Hefei, Anhui P. R. China (Email: jundu@ustc.edu.cn).
	
	
	\newpage
	
	\section{Introduction}
	Due to its promising applications in education and unique academic value pertaining to generalized zero-shot learning \cite{gzsl} and transfer learning \cite{tranfer_survey}, handwritten Chinese character error correction (HCCEC) has attracted rising attention in recent years. It comprises two primary subtasks, namely \textit{assessment} and \textit{correction}. The assessment subtask entails the determination of whether a given handwritten character is spelled wrongly, while the correction subtask involves rectifying the errors by predicting the ideal character that the user intended to write. Throughout the subsequent sections, we represent the wrongly spelled characters as misspelled characters and rightly spelled ones as right characters.
	
\begin{figure*}[htb]
		\centerline{\includegraphics[width=0.65\linewidth]{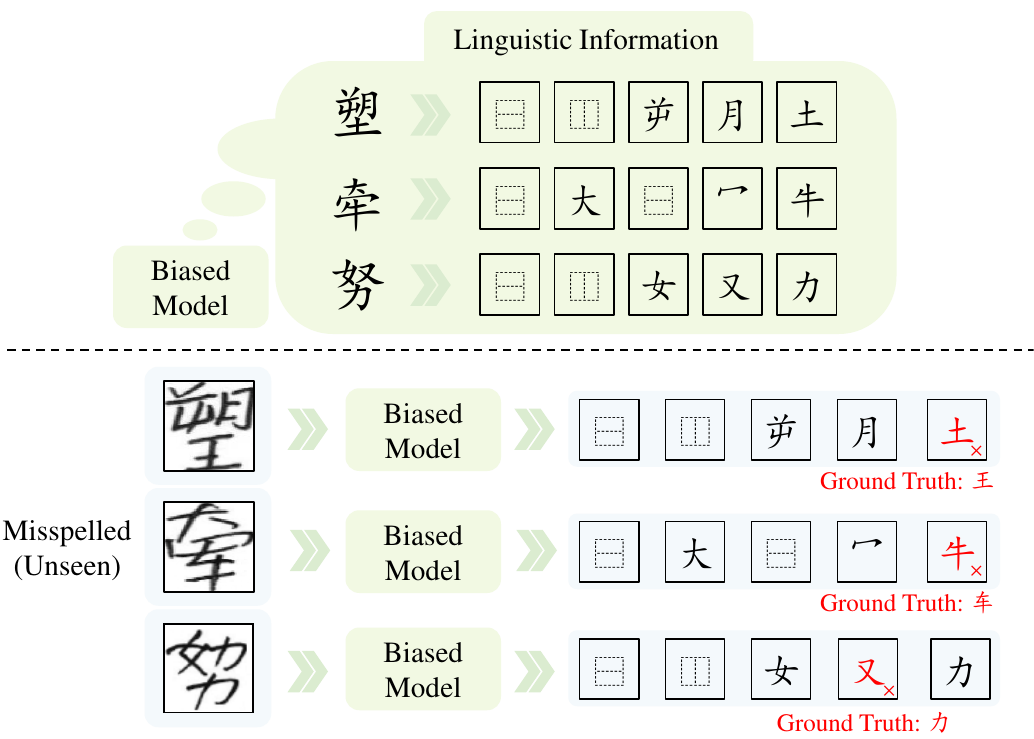}}
		\caption{Top: During training, the biased model implicitly captures and encodes linguistic information inherent in IDS sequences. Bottom: During testing, when presented with a misspelled character (unseen), the biased model wrongly generates an IDS sequence that matches a seen character.}
		\label{img:motivation}
	\end{figure*}

	The utilization of encoder-decoder models has facilitated significant progress in HCCEC. These models decompose a Chinese character into radicals in a sequence \cite{ran} or tree structure \cite{tan} in order to determine the correctness and rectify any possible errors. Unlike other tasks that also involve the decomposition of Chinese characters, such as Chinese character recognition \cite{denseran}, HCCEC requires models to deal with misspelled characters besides right ones. This presents new challenges: 1. Due to the innumerable and unpredictable categories of misspelled characters, the HCCEC test set contains both seen characters and unseen misspelled ones, thus posing a generalized zero-shot learning problem \cite{gzsl}. 2. A misspelled character may differ from its corresponding right character in only one or several radicals, resulting in a high degree of similarity between the two. Consequently, as illustrated in Figure \ref{img:motivation}, a strong bias problem \cite{zero_shot_bias} may arise, where instances of unseen misspelled characters are more likely to be decomposed into an ideographic description sequence (IDS) that matches one of the seen characters.
	
	When generating IDS sequences from character images, models can utilize two types of information: visual information and linguistic information. Visual information signifies the glyph of radicals depicted in the image, while linguistic information implicitly embodies the grammatical rules and correlations between radicals. Existing encoder-decoder methods implicitly model the linguistic information by recursively using the radical predicted from the last decoding time step. The utilization of linguistic information has been extensively explored in other fields, such as scene text recognition \cite{mishra2012scene}. It has demonstrated the capacity to deduce the correct character by leveraging the overall word context when recognizing images with confused  visual cues such as occlusion, blur, and noise \cite{srn,wang2021two,fang2021read}. However, in this paper, we argue that the use of linguistic information is not always advantageous in HCCEC. While linguistic information can be employed to assist the recognition of seen characters with complicated structures or long IDS sequences, it may mislead the decomposition of unseen misspelled characters. Therefore, a good method for HCCEC should mitigate linguistic information for unseen misspelled characters while preserving it for seen characters adaptively.
	
	In this study, we introduce Count, Decode and Fetch (CDF), a novel method for HCCEC. In contrast to existing encoder-decoder methods, CDF exhibits enhanced generalization capabilities toward unseen misspelled characters, owing to its proper utilization of linguistic information. Specifically, CDF comprises three primary components: the counter, the decoder, and the fetcher. Given an image of a Chinese character that can be either misspelled or not, firstly, the counter predicts a counting vector that indicates the number of each radical class. It is worth mentioning that the counter is weakly-supervised \cite{liang2022transcrowd} and requires only annotations of the number of each radical class, which can be easily obtained from the IDS sequences. Subsequently, with the assistance of the counting vector, the decoder generates the IDS sequence step by step. The counting vector is updated according to the output of the decoder at each time step. It provides additional historical alignment information and plays an important role in predicting the current radical. Upon obtaining the IDS sequence, we consult the IDS dictionary under the national standard GB18030-2005 to verify the correctness. If the decomposed IDS sequence is not included, the sample will be identified as a misspelled character. In such cases, the fetcher will predict the ideal character that the user intended to write. Since large-scale misspelled characters are not available due to their rarity and difficulty of collecting \cite{tan}, the fetcher is designed based on the transductive transfer learning strategy \cite{tranfer_survey}, and requires only right characters for training. Several techniques are employed to bridge the gaps between the training and test domains. Moreover, we apply CDF to various encoder-decoder models \cite{ran,tan} and enhance their performance to a certain degree. This demonstrates the effectiveness of the generalization capabilities of our method.
	
	
	The main contributions of this paper are as follows:
	\begin{itemize}
		
		\item
		We highlight the impact of linguistic information on handwritten Chinese character error correction, and propose a novel approach called Count, Decode and Fetch (CDF). We employ a decoder integrated with a counter, which efficiently mitigates linguistic information for unseen misspelled characters while simultaneously preserving it for seen characters.
		
		\item
		We introduce a sophisticated fetcher that utilizes the transductive transfer learning strategy to predict ideal characters. It effectively overcomes the scarcity of misspelled characters and requires only right characters for training.
		
		\item 
		Our CDF can be generalized to various encoder-decoder models and achieve new state-of-the-art results. The experiment results demonstrate the superiority of our method over the previous best method by a large margin, with an impressive 18.7\% improvement in the correct rate and a 10.4\% enhancement in the decomposition accuracy of misspelled characters.
		
	\end{itemize}
	
	\section{Related Work}
	\subsection{Chinese Character Decomposition}
	Chinese character decomposition has been studied for decades due to its technical challenges and great social needs \cite{dai2007chinese}. It is essential for Chinese character recognition \cite{liu2001model,ma2008new}, particularly in the zero-shot/few-shot setting \cite{rcn}, and plays a critical role in tasks such as text image super-resolution \cite{stroke_super_resolution}, character style transfer \cite{huang2020rd} and font generation \cite{zeng2021strokegan}. Based on the decomposition level of a Chinese character, existing methods can be broadly categorized into two approaches: radical-level \cite{ran,hde} and stroke-level \cite{sld,liu2001model,su2003novel}. \cite{radical_1th} employs a recursive hierarchical scheme to separate radicals and proposes hierarchical radical matching to identify characters. Multilabel learning with a residual network architecture is proposed in \cite{radical_2th}. It first defines twenty types of radical structures and then marks each radical with a specific position. Inspired by successful applications of encoder-decoder models in machine translation \cite{translation}, mathematical expression recognition \cite{wap}, document analysis \cite{mtd,semv2,hrdoc}, \cite{ran} introduces an encoder-decoder model to decompose a Chinese character into a sequence of radicals, achieving great success in Chinese character recognition. \cite{fewshotran} further combines deep prototype learning for more robust feature extraction. In \cite{hde}, a hierarchical decomposition embedding method is introduced to represent a Chinese character with a semantic vector. In contrast to radical-level decomposition, stroke-level decomposition treats a Chinese character as a combination of strokes. \cite{stroke_1th} performs a thinning procedure on a character and extracts strokes based on the thinning results. For reliable stroke extraction, mathematical morphology is further proposed in \cite{stroke_morphology} to deal with various shape distortions. Recently, a deep learning based method is proposed in \cite{sld}, which adopts the encoder-decoder architecture and decomposes a Chinese character into a combination of five strokes, including horizontal, vertical, left-falling, right-falling, and turning.

	\subsection{Handwritten Chinese Character Error Correction}
	Handwritten Chinese character error correction (HCCEC) comprises two primary subtasks, namely assessment and correction, which can be sequentially solved. The assessment subtask entails determining whether a given handwritten character is spelled wrongly, and the correction subtask involves rectifying the errors by predicting the ideal character that the user intended to write. \cite{tan} proposes the first model for HCCEC, and adapts several existing models to this task \cite{ran,rcn,hde}. RAN \cite{ran} employs an encoder-decoder framework to decompose Chinese characters into IDS sequences. TAN \cite{tan} further introduces a tree-structured decoder to decompose Chinese characters into radical-tree layouts.  Upon obtaining the IDS sequence or the radical-tree layout, the correctness can be verified by consulting the IDS dictionary. If the decomposed result is not included in the dictionary, the sample will be identified as a misspelled character. In such cases, the decomposed results are utilized to characterize each character, and the most closely matching candidate character is selected as the ideal character. On the other hand, instead of decomposing Chinese characters, RCN \cite{rcn} and HDE \cite{hde} utilize metric-learning techniques to represent a character as a radical-based embedding. In this paper, we focus on the encoder-decoder framework due to its interpretability. 
	
	A recent study \cite{openset} introduces a new open-set text recognition task. It necessitates the model to identify and reject samples of out-of-set characters. From a certain perspective, misspelled characters could be regarded as out-of-set characters. However, HCCEC exhibits notable distinctions in the following aspects: (1) Misspelled characters and their corresponding right ones are highly similar in overall appearance, but differ in fine-grained features. (2) HCCEC incorporates an extra subtask, namely correction, which involves predicting the ideal characters for the misspelled ones.

	\subsection{Objects Counting}
	Counting objects from images \cite{arteta2016counting,levy2015live,zhang2020context} is a fundamental problem in computer vision with a wide range of applications, including public transportation \cite{lengvenis2013application}, physical exercise analysis \cite{soro2019recognition}, and microscopy image analysis \cite{xie2018microscopy}. Traditional methods for this task rely on object detection \cite{leibe2005pedestrian,wang2011automatic,stewart2016end}, where the count value is determined by the number of bounding boxes detected. However, these methods suffer from occlusions in congested regions and require expensive manual annotation. An alternative approach to object counting is to regress a density map \cite{bai2020adaptive,li2018csrnet,baixiang_count1}, where the count value is reflected by integrating the density map. Recently, a counting-aware network \cite{can} is proposed, which employs a multi-scale counting module for handwritten mathematical expression recognition, since the counting results can serve as additional global information to promote recognition accuracy. Besides, a Chinese character recognition method named augmented character profile matching is proposed in \cite{acpm}, which introduces a counting module to perceive the totally numerical values of radicals in order to comprehensively characterize the input image. 

	\section{Preliminaries}
	\begin{figure*}[htb]
		\centerline{\includegraphics[width=1.\linewidth]{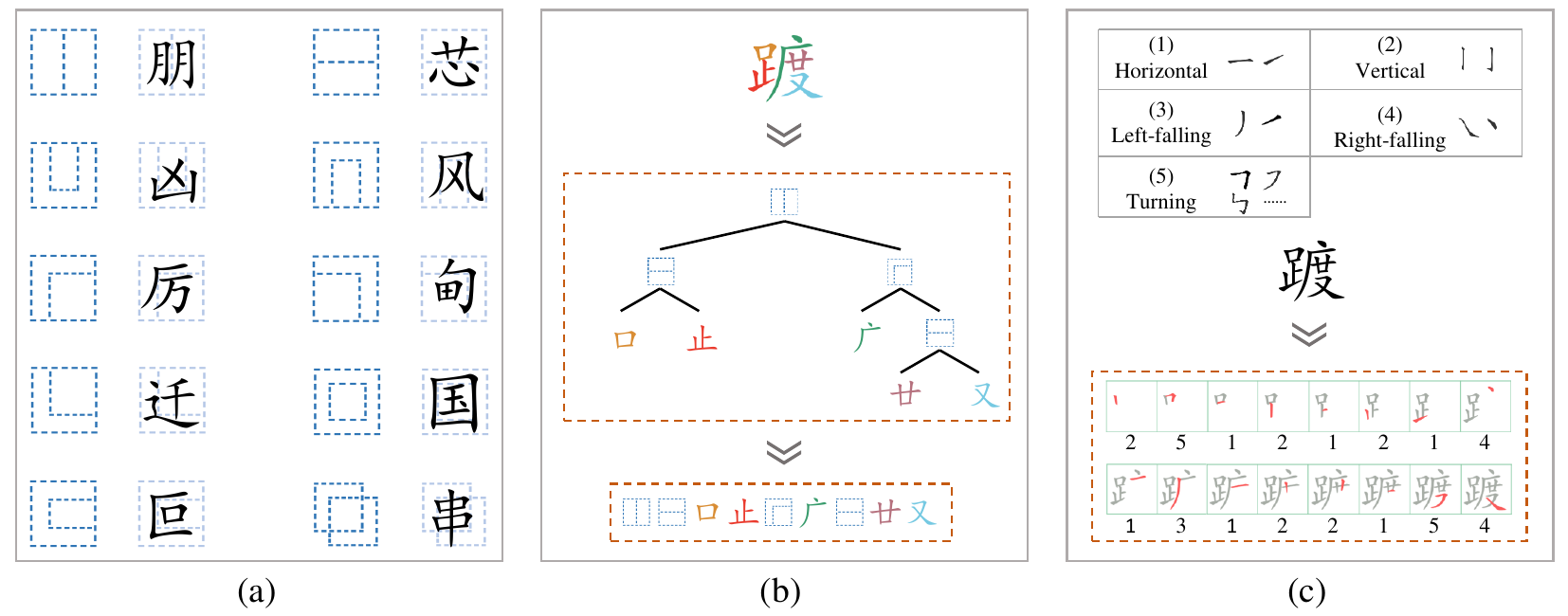}}
		\caption{(a) Graphical representation of ten radical structures. (b) Radical-level decomposition of an example Chinese character. (c) Stroke-level decomposition of an example Chinese character. }
		\label{img:background_decomposition}
	\end{figure*}
	\subsection{Chinese Character Decomposition}
	It is widely acknowledged that Chinese characters can be decomposed at the radical or stroke level. At the radical level, ten different Chinese radical structures are predefined \cite{ran} and we present them in Figure \ref{img:background_decomposition} (a): (1) left-right, (2) above-below, (3) bottom-surround, (4) top-surround, (5) top-left-surround, (6) top-right-surround, (7) bottom-left-surround, (8) full-surround, (9) left-surround, and (10) overlaid. Accordingly, the hierarchical radical structure of a Chinese character can be represented as a tree, as illustrated in Figure \ref{img:background_decomposition} (b). The Chinese character instance is positioned above the top of the tree, and different radicals are denoted with different colors. Symbols on the parent nodes denote radical structures, while symbols on the leaf nodes denote radicals. The IDS sequence can be further obtained by converting from the hierarchical tree structure through a depth-first traversal order.
	
	At the stroke level, a Chinese character can be decomposed into a combination of five fundamental strokes: horizontal, vertical, left-falling, right-falling, and turning. We list them in the upper part of Figure \ref{img:background_decomposition} (c). Accordingly, the Chinese character instance can be decomposed into a sequence of strokes, following a regular order, i.e., left to right, top to bottom, and outside in.
	
	\begin{figure*}[htb]
		\centerline{\includegraphics[width=0.6\linewidth]{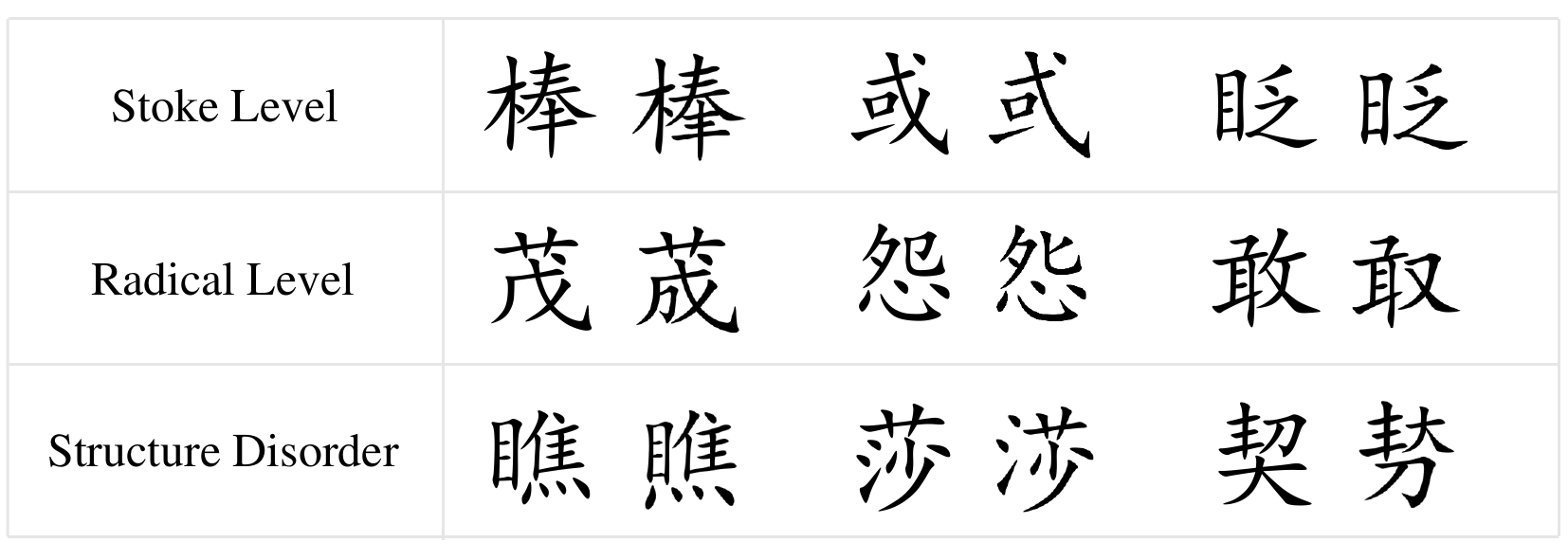}}
		\caption{Examples of three misspelled error types. The right one of each pair is the misspelled character and the left one is the corresponding right character.}
		\label{img:background_misspelled}
		
	\end{figure*}
	
	\subsection{Misspelled Chinese Character}
	
	Misspelled Chinese characters are those that are not included in the national standard GB18030-2005. They are created due to incorrect handwriting by users, who originally intended to write the corresponding right characters that are included in the national standard GB18030-2005. Character errors can be classified into three categories \cite{tan}: stroke-level error, radical-level error, and structure disorder. Figure \ref{img:background_misspelled} presents examples of these misspelled error types. Stroke-level error refers to mistakes involving the addition, deletion, or misuse of one stroke. Radical-level error involves analogous errors with one or more radicals. And structure disorder refers to correct radicals that are composed in the wrong structural order. 
	
	Considering that the addition or deletion of strokes can be transformed into the difference between similar radicals, methods for HCCEC typically model Chinese characters at the radical level. Stroke-level decomposition is not suitable for HCCEC for two reasons: (1) it suffers from a one-to-many problem \cite{sld}, where different characters may share the same stroke sequence, and (2) radicals are easier to identify and locate than strokes, which aligns better with human intuition.
	
	\begin{figure*}[htb]
		\centerline{\includegraphics[width=0.8\linewidth]{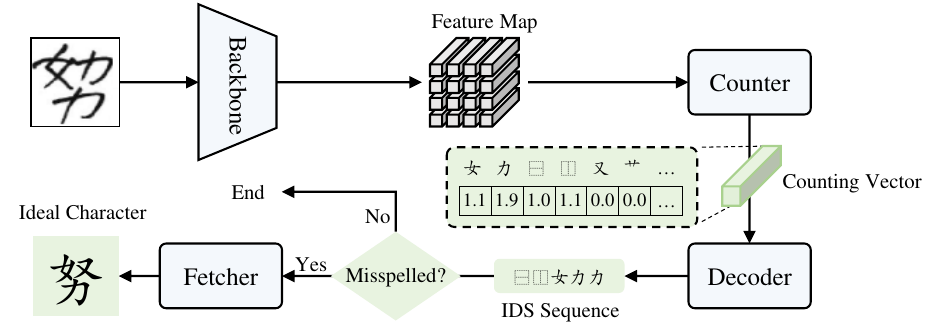}}
		\caption{The flow chart of the proposed CDF, which consists of three modules: Counter, Decoder and Fetcher.}
		\label{img:pipeline}
	\end{figure*}
	
	\section{Method}
	The overall pipeline of our system is shown in Figure \ref{img:pipeline}, which consists of a backbone, a counter, a decoder, and a fetcher. Given an image $I$ of handwritten Chinese characters, DenseNet \cite{densenet} is applied as the backbone following previous work \cite{ran,tan}. It extracts the feature map $\boldsymbol{F} \in \mathbb{R}^{L \times C}$ where $L = H \times W$. The counter further utilizes the feature map $\boldsymbol{F}$ to predict the number of each radical class and generates the counting vector $\boldsymbol{\mathcal{C}}$. Subsequently, the counting vector $\boldsymbol{\mathcal{C}}$ and the feature map $\boldsymbol{F}$ are fed into the decoder to decompose the given character into an IDS sequence, which allows for accessing the correctness by looking up the IDS dictionary. In cases of a misspelling, the fetcher will provide the ideal character to the user based on the feature map $\boldsymbol{F}$ and the decomposed IDS sequence. In the following subsections, three main components in our system, namely the counter, the decoder and the fetcher, will be elaborated.

	\subsection{Counter}
	
	\begin{figure*}[htb]
		\centerline{\includegraphics[width=0.7\linewidth]{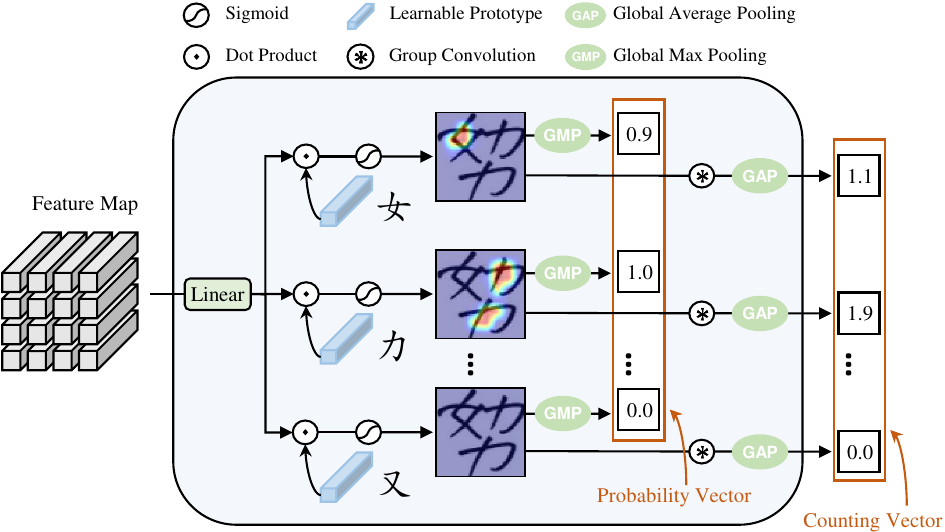}}
		\caption{The illustration of the counter. It takes the feature map as input and predicts the counting vector, which denotes the number of each radical class.}
		\label{count}
	\end{figure*}
	The counter takes the feature map $\boldsymbol{F}$ as input, and generates a counting vector $\boldsymbol{\mathcal{C}}$ that denotes the number of each radical class. Specifically, the counter firstly predicts a probability vector $\boldsymbol{\mathcal{P}}$ that denotes whether each radical exists, then estimates the number of existing radicals and outputs the counting vector $\boldsymbol{\mathcal{C}}$. Notably, the counter predicts the number of each radical class independently, making $\boldsymbol{\mathcal{C}}$ free of linguistic information.
	
	Considering samples of the same radical class bear a striking resemblance in the glyph, we learn a prototype $\boldsymbol{r}_n \in \mathbb{R}^{d}$ for $n^{\text{ th}}$ radical class. Then we compute the energy map $\boldsymbol{E}_n$ using the prototype $\boldsymbol{r}_n$ and the feature map $\boldsymbol{F}$ as follows:
	\begin{equation}
		\boldsymbol{E}_n^* = {\rm sigmoid}(\boldsymbol{F}\boldsymbol{W}_{\rm r}\boldsymbol{r}_n)
	\end{equation}
	Here,  $\boldsymbol{W}_{\rm r} \in \mathbb{R}^{C \times d}$ is a learnable matrix to measure the compatibility between each prototype of radicals and each region of the feature map. Then the energy map $\boldsymbol{E}_n^* \in \mathbb{R}^{L}$ is reshaped to $\boldsymbol{E}_n \in \mathbb{R}^{H \times W}$(recall that $L=H \times W$). Assuming that there are a total of $N$ radicals in the vocabulary, the $n^{\text{th}}$ element of the probability vector $\boldsymbol{\mathcal{P}}  \in \mathbb{R}^{N}$ denotes the probability that $n^{\text{th}}$ radical exists. $\mathcal{P}_n$ is obtained by taking the maximum value of the energy map $\boldsymbol{E}_n$:
	\begin{equation}
		\mathcal{P}_n = \mathop{\max}_{x,y} E_{n,x,y}
		\label{equ:p_n}
	\end{equation}
	
	Afterward, counting each radical class can be achieved by identifying peaks on the energy map. Instead of using complicated post-processing algorithms, we employ a lightweight group convolution layer to obtain the counting vector:
	\begin{equation}
		\mathcal{C}_n = {\rm GAP}(\boldsymbol{Q}_n \circledast \boldsymbol{E}_n)
		\label{equ:c_n}
	\end{equation}
	where $\mathcal{C}_n$ is the $n^{\text{th}}$ element of the counting vector $ \boldsymbol{\mathcal{C}} \in \mathbb{R}^{N} $ and indicates the predicted number of $n^{\text{th}}$ radical class. GAP refers to global average pooling. $\circledast$ denotes a convolution operation and $\boldsymbol{Q}_n\in \mathbb{R}^{k_w \times k_h}$ is the filter. $k_w$ and $k_h$ refer to the width and height of the kernel respectively. It is noteworthy that all elements of $\boldsymbol{\mathcal{C}}$ can be obtained in parallel through one convolution layer by setting the number of filter groups \cite{group_conv} to $N$ (the total number of radical classes). This reduces the number of parameters and ensures that the count value for each class is predicted separately, thus free of linguistic information.
	
	The loss function for the counter is defined as follows:
	\begin{equation}
		L_{\text{c}} = \frac{1}{N}L_{\text{cls}}(\widetilde{\boldsymbol{\mathcal{P}}}, \boldsymbol{\mathcal{P}}) + \frac{1}{N^{\prime}}\sum_{n=1}^{N}{\rm I}_n{\rm Smooth}_{L_1}
		(\widetilde{\mathcal{C}}_n - \mathcal{C}_n)
		\label{equ:L_c}
	\end{equation}
    in which
    \begin{equation}
    	L_{\text{cls}}(\boldsymbol{\widetilde{\mathcal{P}}}, \boldsymbol{\mathcal{P}}) = \sum_{n=1}^{N}-(\widetilde{\mathcal{P}}_n\log(\mathcal{P}_n) + 
    	(1 - \widetilde{\mathcal{P}}_n)\log(1 - \mathcal{P}_n))
    \end{equation}
	Here, ${\rm Smooth}_{L_1}$ denotes a smooth $L_1$ regression loss \cite{fastrcnn}. ${\rm I}_n$ is an indicator that evaluates to $1$ when $\mathcal{P}_n > 0.5$ and $0$ otherwise. $N' = \sum_{n=1}^{N}{\rm I}_n$. $\widetilde{\mathcal{P}}_n$ and $\widetilde{\mathcal{C}}_n$ represent the corresponding ground truth of $\mathcal{P}_n$ and $\mathcal{C}_n$, respectively.
	
	\begin{figure*}[htb]
		\centerline{\includegraphics[width=0.45\linewidth]{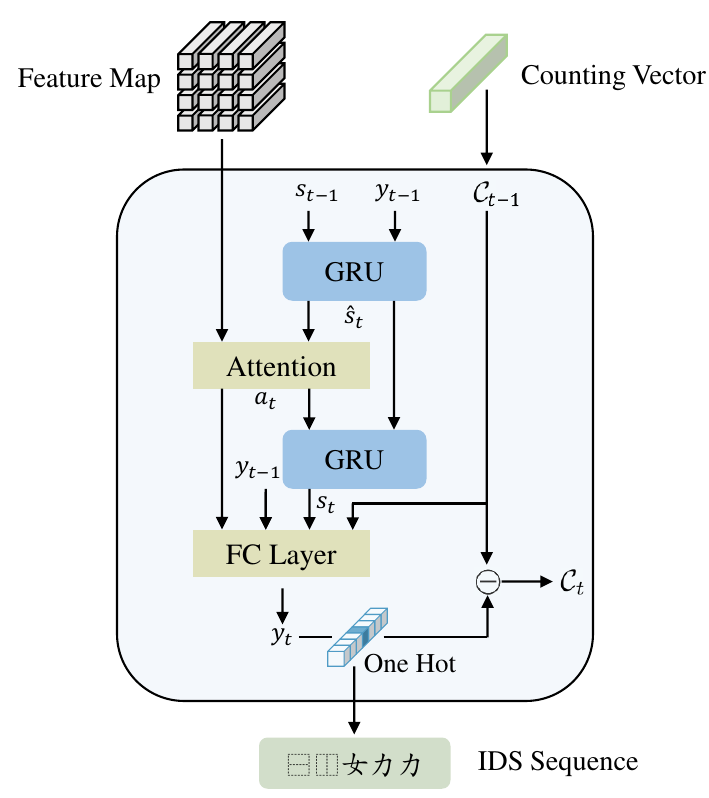}}
		\caption{The illustration of the decoder. It takes the feature map and the counting vector as input, and generates the IDS sequence step by step.}
		\label{img:decoder}
	\end{figure*}
	
	\subsection{Decoder \label{sec:decoder}}
	
	The decoder is designed to leverage linguistic information adaptively. It takes the feature map $\boldsymbol{F}$ and the counting vector $\boldsymbol{\mathcal{C}}$ as input, and generates the IDS sequence step-by-step. At decoding step $t$, the current query vector $\boldsymbol{\hat{s}}_t \in \mathbb{R}^{d_s}$ is computed from the previous hidden state $\boldsymbol{s}_{t-1}$ and the embedding of the previous radical $\boldsymbol{v}_{t-1}$ through the first GRU layer:
	\begin{equation}
		\boldsymbol{v}_{t-1} = {\rm Emb}(y_{t-1})
	\end{equation}
	\begin{equation}
		\boldsymbol{\hat{s}}_t = {\rm GRU}(\boldsymbol{v}_{t-1}, \boldsymbol{s}_{t-1})
	\end{equation}
	
	Then we employ the coverage-based attention mechanism $f_{\rm att}$ to obtain the current context vector $\boldsymbol{a}_t$, which involves the related part of the image. The feature map $\boldsymbol{F}$ assumes the role of both the key and the value, while $\boldsymbol{\hat{s}}_t$ functions as the query:
	\begin{equation}
		\boldsymbol{a}_t = f_{ \rm att}(\boldsymbol{\hat{s}}_t,\boldsymbol{F})
	\end{equation}
	where $f_{\rm att}$ is designed as follows:
	\begin{equation}
		\boldsymbol{H} = {\rm conv} (\sum_{k=1}^{t-1} \boldsymbol{\alpha}_k)
	\end{equation}
	\begin{equation}
		e_{t,i} = \boldsymbol{w}^{\top } \tanh (\boldsymbol{W}_{\rm att}\boldsymbol{\hat{s}}_t + \boldsymbol{U}_{\rm att}\boldsymbol{F}_i + \boldsymbol{W}_{\rm h}\boldsymbol{H}_i)
	\end{equation}
	\begin{equation}
		\alpha_{t,i} = \exp (e_{t,i} ) / \sum_{k=1}^L\exp (e_{t, k})
	\end{equation}
	\begin{equation}
		\boldsymbol{a}_t = \sum_{i=1}^L \alpha_{t,i} \boldsymbol{F}_i
	\end{equation}
	Here, the coverage vector $\boldsymbol{H} \in \mathbb{R}^{L \times d_h}$ is used to address the lack of coverage \cite{coverage_vector}. $\boldsymbol{F}_i$ and $\boldsymbol{H}_i$ refer to the $i^{\text{th}}$ element of $\boldsymbol{F}$ and $\boldsymbol{H}$. Attention coefficients $\alpha_{t,i} \in \mathbb{R}$ are obtained by feeding $e_{t,i}$ into a softmax function. The context vector $\boldsymbol{a}_t \in \mathbb{R}^{d_a}$ is computed by the weighted summation of the elements of the feature map. $\boldsymbol{W}_{\rm att}$, $\boldsymbol{U}_{\rm att}$, $\boldsymbol{W}_{\rm h}$, and $\boldsymbol{w}$ are trainable weights.
	
	The second GRU layer is utilized to calculate the current hidden state $s_t$:
	\begin{equation}
		\boldsymbol{s}_t = {\rm GRU}(\boldsymbol{a}_t, \boldsymbol{\hat{s}}_t)
	\end{equation}
	
	The current radical feature $\boldsymbol{g}_t \in \mathbb{R}^{d_g}$ is obtained with the hidden state $\boldsymbol{s}_t$, the context vector $\boldsymbol{a}_t$, and the embedding of the last radical $\boldsymbol{v}_{t-1}$:
	\begin{equation}
		\boldsymbol{g}_t = \boldsymbol{W}_{\rm v}\boldsymbol{v}_{t-1} + \boldsymbol{W}_{\rm s}\boldsymbol{s}_t + \boldsymbol{W}_{\rm a}\boldsymbol{a}_t
		\label{equ:g_t}
	\end{equation}
	where $\boldsymbol{W}_{\rm v}$, $\boldsymbol{W}_{\rm s}$, and $\boldsymbol{W}_{\rm a}$ are trainable weights.
	
	In most previous methods \cite{ran,tan}, the prediction of $y_t$ relies solely on $\boldsymbol{g}_t$. We further employ a variable vector $\boldsymbol{\mathcal{C}}_{t-1} \in \mathbb{R}^{N}$ to obtain the current radical:
	\begin{equation}
		\boldsymbol{p}_t = {\rm softmax} (\boldsymbol{W}_{\rm p}\varphi(\boldsymbol{g}_t + \boldsymbol{W}_{\rm c}\boldsymbol{\mathcal{C}}_{t-1}))
		\label{equ:p_yt}
	\end{equation}
	\begin{equation}
		y_t = \mathop{\arg\max} \boldsymbol{p}_t
	\end{equation}
	\begin{equation}
		\boldsymbol{\mathcal{C}}_{t} = {\rm relu}(\boldsymbol{\mathcal{C}}_{t-1} - {\rm onehot} (y_t))
		\label{equ:c_t+1}
	\end{equation}
	Here,  $\boldsymbol{\mathcal{C}}_t$ is initialized by the counting vector $\boldsymbol{\mathcal{C}}$ (i.e., $\boldsymbol{\mathcal{C}}_0 = \boldsymbol{\mathcal{C}}$), and it is updated according to $\boldsymbol{y}_t$ at each time step. $\varphi$ denotes a maxout activation function. $\boldsymbol{W}_{\rm p}$ and $\boldsymbol{W}_{\rm c}$ are trainable weights. Variable $\boldsymbol{\mathcal{C}}_t$ contributes to the mitigation of linguistic information for misspelled characters, since it incorporates the past alignment information. For right characters, $\boldsymbol{\mathcal{C}}_t$ is also beneficial as it provides additional counting information.
	
	A cross entropy loss $L_{\text{d}}$ is used as the training objective of the decoder:
	\begin{equation}
		L_{\text{d}} = -\frac{1}{T}\sum_{t=1}^T\boldsymbol{\widetilde{p}}_t\log \boldsymbol{p}_t
		\label{equ:L_d}
	\end{equation}
	where $\boldsymbol{p}_t$ is computed in Eq. \ref{equ:p_yt}, and $\boldsymbol{\widetilde{p}}_t$ denotes its corresponding ground truth.
	
	\begin{figure*}[htb]
		\centerline{\includegraphics[width=0.5\linewidth]{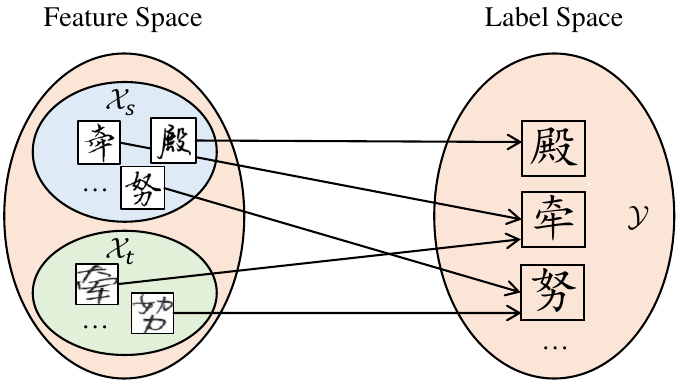}}
		\caption{The formulation of the task of ideal character prediction. $\mathcal{X}_s$ and $\mathcal{X}_t$ denote the feature spaces that involve right character images and misspelled ones, respectively. $\mathcal{Y}$ denotes the corresponding label space that consists of the right character classes.}
		\label{img:fetcher}
	\end{figure*}
	
	\begin{figure*}[htb]
		\centerline{\includegraphics[width=0.47\linewidth]{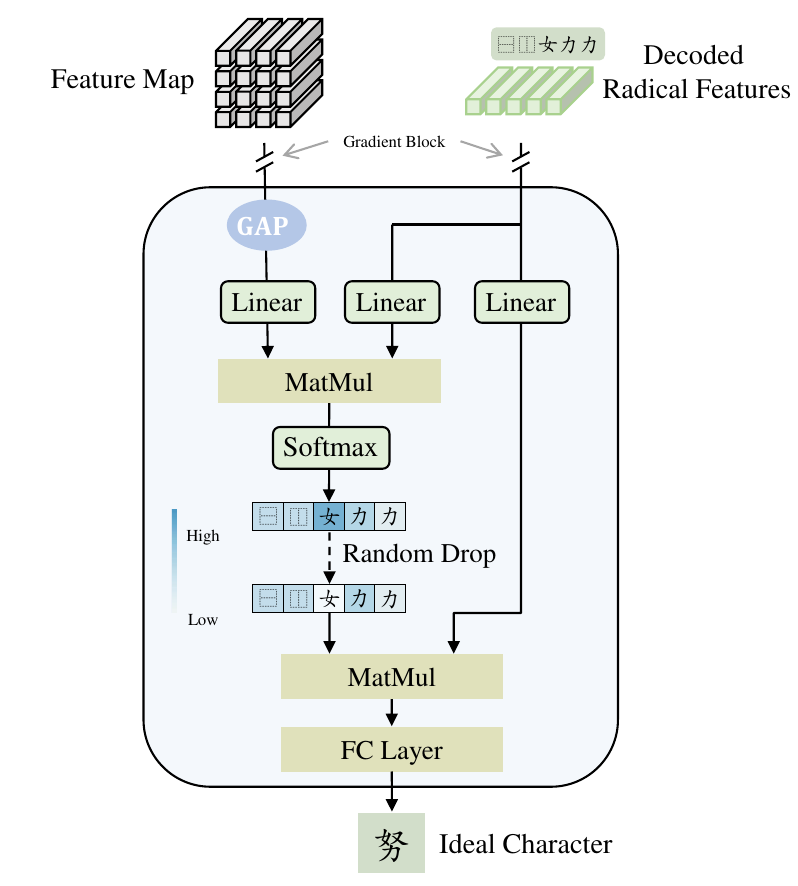}}
		\caption{The illustration of the fetcher. During training, it takes right character samples as input and predicts the corresponding character classes. During testing, it takes misspelled character samples as input and predicts the ideal characters.}
		\label{img:fetcher_network}
	\end{figure*}

	\subsection{Fetcher}
	To correct errors of a misspelling, the ideal character (the character that the user originally intended to write) needs to be predicted. However, as mentioned above, misspelled characters are rare and difficult to collect, resulting in limited training data. Therefore, we employ transductive transfer learning \cite{tranfer_survey} strategy and design a fetcher that requires only right characters for training.
	
	Firstly, we formulate the task of ideal character prediction. As shown in Figure \ref{img:fetcher}, the feature space $\mathcal{X}_s$ in the source domain involves only right character images, while $\mathcal{X}_t$ in the target domain involves only misspelled ones. Although $\mathcal{X}_s \neq \mathcal{X}_t$, the source domain and the target domain share the same label space $\mathcal{Y}$ that consists of right character classes. We aim at adapting the fetcher trained in the source domain for use in the target domain. A major problem is how to reduce the difference between the distributions of source and target domain data. Thanks to the hierarchical structure of Chinese characters, the IDS sequence can play a role to bridge the domain gap.
	
	The radical feature $\boldsymbol{g}_t$ in Equation \ref{equ:g_t} preserves detailed geometric properties of the character. We employ the attention mechanism to select radicals that contain strikingly glyphic information. As shown in Figure \ref{img:fetcher_network}, the query comes from the feature map $\boldsymbol{F}$, serving as additional global information. Both the key and value come from the radical features. Assuming that $\boldsymbol{G} = [\boldsymbol{g}_1;\boldsymbol{g}_2;...;\boldsymbol{g}_T] \in \mathbb{R}^{d_g \times T}$ denotes the concatenation of radical features, the energy of radicals $\boldsymbol{b} \in \mathbb{R}^{T}$ is computed as follows:
	\begin{equation}
		\boldsymbol{Q} = \boldsymbol{U}_{\rm q}{\rm GAP}(\boldsymbol{F})
	\end{equation}
	\begin{equation}
		\boldsymbol{K} = \boldsymbol{U}_k\boldsymbol{G}
	\end{equation}
	\begin{equation}
		\boldsymbol{b} = \frac{\boldsymbol{K}^{\top}\boldsymbol{Q}}{\sqrt{d_{k}}}
	\end{equation}
	Here, GAP denotes the global average pooling. $\boldsymbol{Q} \in \mathbb{R}^{d_k}$ and $\boldsymbol{K} \in \mathbb{R}^{d_k \times T}$ denote the query and key, respectively. $\boldsymbol{U}_{\rm q} \in \mathbb{R}^{d_{k} \times C}$ and $\boldsymbol{U}_{\rm k} \in \mathbb{R}^{d_{k} \times d_g}$ are trainable weights. Considering a misspelled character usually differs from its ideal character in a few radicals, we randomly drop the radicals after feeding $\boldsymbol{b}$ into a softmax function during training:
	\begin{equation}
		\boldsymbol{\beta} = {\rm RandomDrop}({\rm softmax}(\boldsymbol{b}))
		\label{equ:beta}
	\end{equation}
	where $\boldsymbol{\beta} \in \mathbb{R}^{T}$ indicates the attention weights. RandomDrop is implemented by zeroing some of the elements with probability $p$ using samples from a Bernoulli distribution.
	
	Afterward, the character vector $\boldsymbol{m} \in \mathbb{R}^{d_m}$ is obtained by weighted summation of transformed radical features, and the probability distribution $\boldsymbol{p}^{\text{fet}}$ can be further computed through a linear transformation and a softmax function:
	\begin{equation}
		\boldsymbol{m} = \sum^T_{t=1}\beta_t\boldsymbol{U}_{\rm v}\boldsymbol{g}_t
	\end{equation}
	\begin{equation}
		\boldsymbol{p}^{\rm fet} = {\rm softmax}(\boldsymbol{U}_{\rm f}\boldsymbol{m})
	\end{equation}
	Here, $\boldsymbol{U}_{\rm f} \in \mathbb{R}^{M\times d_m}$ and $\boldsymbol{U}_{\rm v} \in \mathbb{R}^{d_m \times d_g}$ are trainable weights. $M$ denotes the total number of categories of right characters. Suppose the one-hot label of the ideal character is $\boldsymbol{q}^{\rm fet}$, the loss function for the fetcher is defined as follows:
	\begin{equation}
		L_{\text f} = - \boldsymbol{q}^{\rm fet}\log(\boldsymbol{p}^{\rm fet})
		\label{equ:L_f}
	\end{equation}
	
	It is noteworthy that the gradient flow is blocked between the fetcher and other modules. By doing this, $\boldsymbol{g_t}$ and $\boldsymbol{F}$ are protected against overfitting to the source domain.
	
	\begin{CJK}{UTF8}{gkai}	
		\section{Implementation Details}
		\begin{figure*}[htb]
			\centerline{\includegraphics[width=0.5\linewidth]{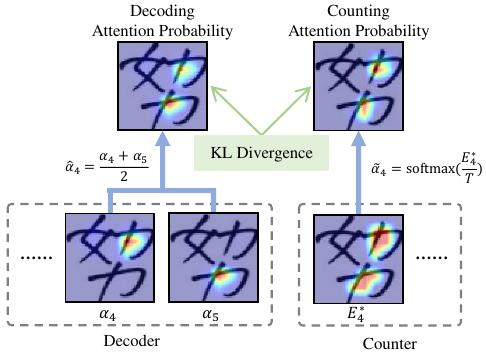}}
			\caption{Attention regularization between the decoder and the counter with regards to the radical ``力''. }
			\label{img:attention_regular}
		\end{figure*}
		\subsection{Attention Regularization}
		There exists a correlation between decoding attention probabilities and counting attention probabilities. This can be illustrated through the example depicted in Figure \ref{img:attention_regular}. The decoder predicts the radical ``力'' at time steps 4 and 5, based on which the decoding attention probability can be generated, while the counter computes the counting attention probability for the same radical ``力''. Since they both attend to the same radicals, it can be inferred that the decoding attention probability should be similar to the counting attention probability.

		To implement the regularization between count attention and decode attention, we undertake the following steps.
		Firstly, assuming that the decoder predicts the $l^{\text{th}}$ radical class $\xi$ at time steps $t_1,...,t_s$, we obtain the decoding attention probability $\boldsymbol{\hat{\alpha}}_l$ by taking the mean value in each region of the corresponding attention maps $\{ \boldsymbol{\alpha}_{t_1},...,\boldsymbol{\alpha}_{t_s}\}$ (e.g., for the $4^{\text{th}}$ radical class ``力'' illustrated in Figure \ref{img:attention_regular}, we get $\boldsymbol{\hat{\alpha}}_4 = \frac{\boldsymbol{\alpha}_4 + \boldsymbol{\alpha}_5}{2}$). 
		Secondly, the counting attention probability $\boldsymbol{\widetilde{\alpha}}_l$ can be obtained by feeding the energy map $\boldsymbol{E}_l^*$ into a softmax function with a hyper-parameter $T$
		(e.g., $\boldsymbol{\widetilde{\alpha}}_4 = {\rm softmax}(\frac{\boldsymbol{E}_4^*}{T})$, and we set $T = 0.2$ here). 
		Finally, supposing that there are $\mathcal{L}$ radicals of different classes in the decomposed IDS sequence, we employ the Kullback-Leibler (KL) divergence between  $\boldsymbol{\widetilde{\alpha}}_l$ and $\boldsymbol{\hat{\alpha}}_l$ as the regularization function:
	\end{CJK}
	
	\begin{equation}
		L_{\text r} = \frac{1}{\mathcal{L}} \sum_{l=1}^{\mathcal{L}} \boldsymbol{\hat{\alpha}}_l \log (\frac{\boldsymbol{\hat{\alpha}}_l}{\boldsymbol{\widetilde{\alpha}}_l})
		\label{equ:L_r}
	\end{equation}
	
	\subsection{Training}
	The overall model is trained end-to-end. The training objective of our model is to minimize the counting loss (Eq. (\ref{equ:L_c})), recognition loss (Eq. (\ref{equ:L_d}),  fetching loss (Eq. (\ref{equ:L_f})) and attention regularization loss (Eq. (\ref{equ:L_r})). The objective function for optimization is shown as follows:
	\begin{equation}
		O = \lambda_1 L_{\text c} + \lambda_2\ L_{\text d} + \lambda_3 L_{\text f} + \lambda_4 L_{\text r}
	\end{equation}
	In our experiments, we set $\lambda_1 = \lambda_2 =\lambda_3 = 1$ to reflect the fact	that the counter, decoder and fetcher are equally important. We set $\lambda_4 = 0.5$ for the attention regularization loss.
	
	To be fairly comparable with previous methods, we use the same DenseNet \cite{densenet} encoder employed in the baseline model RAN \cite{ran} and TAN \cite{tan}. The DenseNet mainly consists of three dense blocks and two transition layers. Each dense block contains 22 bottlenecks. The growth rate of each bottleneck is set to 24. In the counter, the radical prototype dimension $d$ is set to $256$. The width and height of the kernel, $k_w$ and $k_h$, are both set to $8$. In the fetcher, the length of the character vector $d_m$ is set to $256$. The probability $p$ for RandomDrop is set to $0.3$. Other settings in the decoder are consistent with those in the RAN and TAN.
	
	During training, we employ the adadelta algorithm \cite{adadelta} for optimization, with the following hyper-parameters: $ \rho = 0.95$ and $\epsilon = 10^{-6}$. The whole framework is implemented
	using PyTorch. We use a single Nvidia Tesla	V100 with 32GB RAM to train our model with batch size 96.
	
	\subsection{Testing}
	During the testing phase, we introduce a re-weight inference mechanism in the decoder. Specifically, we utilize the counting vector $\boldsymbol{\mathcal{C}}_{t - 1}$ to recalculate the probability distribution of the $y_t$ as follows:
	\begin{equation}
		\boldsymbol{p}^*_t = \boldsymbol{p}_t \odot \tanh (\boldsymbol{\mathcal{C}}_{t - 1} + \delta)
		\label{eq:reweight}
	\end{equation}
	Here, $\odot$ denotes the Hadamard product. $\delta$ is a small constant and we set $\delta = 0.7$ in the experiments. $\delta$ is added to $\boldsymbol{\mathcal{C}}_{t-1}$ through broadcasting to prevent elements of $\boldsymbol{p}^*_t$ from approaching 0, thus enhancing the robustness of the re-weight inference mechanism. To ensure a fair comparison, probability-based decoding proposed in \cite{tan} is also employed in our experiments.

	\section{Experiments}
	\begin{figure*}[htb]
		\centerline{\includegraphics[width=0.55\linewidth]{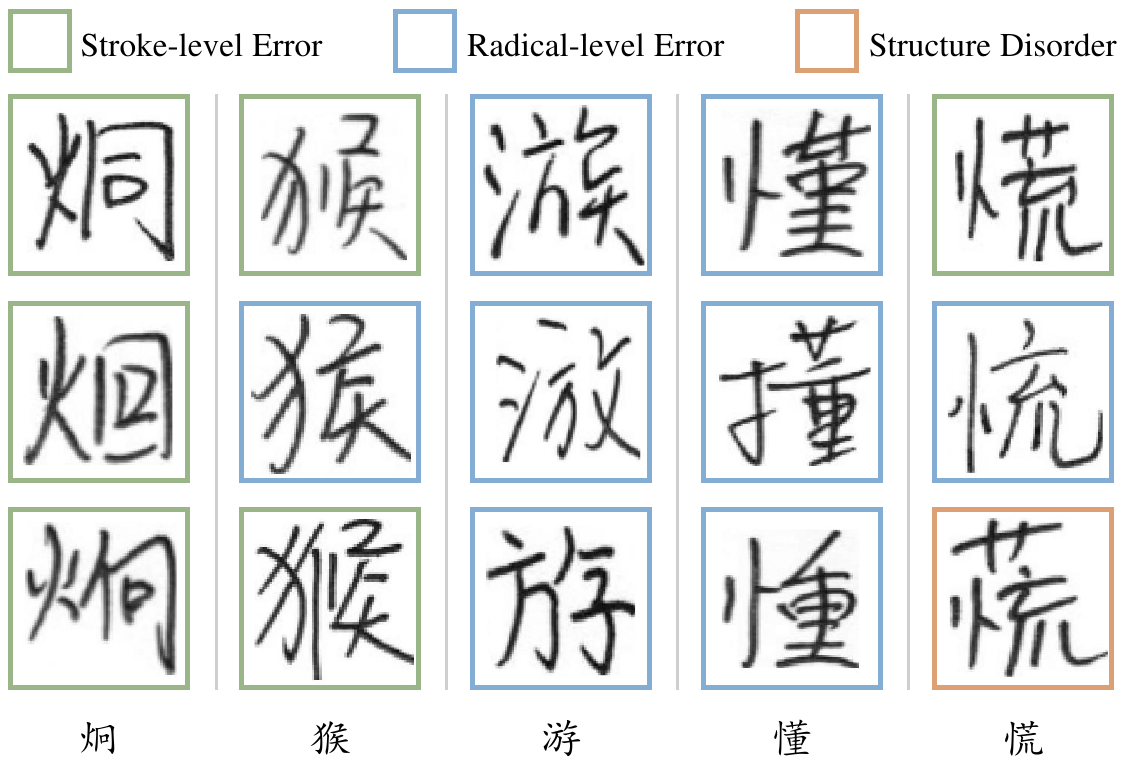}}
		\caption{Example images of three misspelled error types. The lowermost row exhibits the ideal characters for each respective column.}
		\label{img:dataset}
	\end{figure*}
	\subsection{Dataset}
	There are several datasets of Chinese characters such as Chinese Text in the Wild (CTW) \cite{ctw}, HWDB1.0-1.1  \cite{hwdb} and ICDAR 2013 \cite{icdar2013}. Since we focus on error correction, the misspelled handwritten Chinese characters introduced in \cite{tan} are utilized in this paper. The training set of the dataset comprises 5,000 categories of right characters, while the test set comprises 2,000 categories of seen right characters and 570 categories of unseen misspelled characters. And 500 categories of unseen right characters are used as a separate validation set. In total, it includes 400,000 right character samples and 11,400 misspelled character samples. Among these misspelled samples, the proportions of stroke-level errors, radical-level errors and structure disorders are 41.1\%, 56.1\%, and 2.8\%, respectively. We show some example images of three kinds of misspelled characters in Figure \ref{img:dataset}.
	
	\subsection{Evaluation Metrics}
	\textbf{$\bm{F_1}$-score} is a metric for the assessment subtask. It measures the performance of a model in determining whether a given sample is misspelled or correctly spelled.
	
	\textbf{Decomposition Accuracy} (DACC) is a more refined measure for the assessment subtask. It evaluates the ability of a model to decompose the given sample into the corresponding IDS sequence.
	
	\textbf{Ideal Accuracy} (IACC) is a metric for the correction subtask. It measures the ability of a model to predict the ideal characters for misspelled samples. Following previous work, we allow top-$k$ candidate predictions for each sample, and we set $k=5$ by default.
	
	\textbf{Correction Rate} (CR) is an overall metric for both the decomposition and correction subtasks. It requires the model not only to decompose the given sample correctly, but also to predict the ideal character correctly.
	
	\textbf{Mean Absolute/Squared Error} (MAE/MSE) are widely used metrics in multi-class object counting tasks \cite{multi_class_counting}. In this study, they are employed to measure the performance of radical counting.

	\subsection{Comparison with State-Of-The-Art Methods}
	
	\begin{table*}[]
		\caption{Comparison with previous state-of-the-art methods (in \%).  The Misspelled Set is divided into subsets based on error type, namely Stroke, Radical, and Structure. P and R denote precision and recall, respectively.}
		\label{table:sota}
		\centering
		\resizebox{1.0\columnwidth}{!}{%
			\begin{tabular}{@{}l|ccccccccccc|cccc@{}}
				\toprule
				\multirow{3}{*}{Method} & \multicolumn{11}{c|}{Misspelled Set} & \multicolumn{4}{c}{Right Set} \\ \cmidrule(l){2-16} 
				& \multicolumn{2}{c|}{Stroke} & \multicolumn{2}{c|}{Radcal} & \multicolumn{2}{c|}{Structure} & \multirow{2}{*}{DACC} & \multirow{2}{*}{CR} & \multirow{2}{*}{P} & \multirow{2}{*}{R} & \multirow{2}{*}{F1} & \multirow{2}{*}{DACC} & \multirow{2}{*}{P} & \multirow{2}{*}{R} & \multirow{2}{*}{F1} \\ \cmidrule(lr){2-7}
				& DACC & \multicolumn{1}{c|}{CR} & DACC & \multicolumn{1}{c|}{CR} & DACC & \multicolumn{1}{c|}{CR} &  &  &  &  &  &  &  &  &  \\ \midrule
				RCN \cite{rcn} & 29.4 & \multicolumn{1}{c|}{17.8} & 54.2 & \multicolumn{1}{c|}{26.3} & 0.0 & \multicolumn{1}{c|}{0.0} & 42.4 & 22.1 & 97.2 & 45.2 & 61.7 & 92.4 & 86.5 & 99.7 & 92.6 \\
				HDE \cite{hde} & 32.7 & \multicolumn{1}{c|}{27.6} & 59.0 & \multicolumn{1}{c|}{42.0} & 54.9 & \multicolumn{1}{c|}{51.8} & 47.2 & 32.9 & 96.7 & 50.5 & 66.3 & 92.1 & 87.6 & 99.5 & 93.2 \\ \midrule
				RAN (baseline) \cite{ran} & 43.4 & \multicolumn{1}{c|}{31.1} & 67.2 & \multicolumn{1}{c|}{39.3} & 60.9 & \multicolumn{1}{c|}{38.8} & 57.3 & 35.9 & \textbf{98.2} & 59.2 & 73.9 & 94.5 & 89.6 & 99.7 & 94.4 \\
				CDF-RAN (ours) & \textbf{58.0} & \multicolumn{1}{c|}{\textbf{49.6}} & \textbf{76.1} & \multicolumn{1}{c|}{\textbf{63.0}} & \textbf{65.3} & \multicolumn{1}{c|}{\textbf{60.0}} & \textbf{68.4} & \textbf{57.4} & 97.6 & \textbf{70.2} & \textbf{81.7} & 94.6 & 92.1 & 99.5 & 95.7 \\ \midrule
				TAN (baseline) \cite{tan} & 44.2 & \multicolumn{1}{c|}{33.3} & 68.0 & \multicolumn{1}{c|}{42.0} & 62.2 & \multicolumn{1}{c|}{51.8} & 58.0 & 38.7 & 98.0 & 60.0 & 74.4 & 94.6 & 89.7 & \textbf{99.7} & 94.4 \\
				CDF-TAN (ours) & 54.0 & \multicolumn{1}{c|}{46.5} & 72.8 & \multicolumn{1}{c|}{57.9} & 64.3 & \multicolumn{1}{c|}{59.6} & 64.8 & 53.3 & 97.8 & 65.2 & 78.2 & \textbf{94.7} & \textbf{92.3} & 99.6 & \textbf{95.8} \\ \bottomrule
			\end{tabular}%
		}
	\end{table*}
	
	\begin{table}[]
		\caption{Comparison on ideal character prediction with previous algorithms with metric Ideal Accuracy (in \%).}
		\label{tab:ideal}
		\centering
		\resizebox{0.5\columnwidth}{!}{%
			\begin{tabular}{@{}l|ccccc@{}}
				\toprule
				Method & Top 1 & Top 2 & Top 3 & Top 4 & Top 5 \\ \midrule
				Edit-Distance \cite{ran} & 33.0 & 36.8 & 39.6 & 43.3 & 44.5 \\
				Probability-Embedding \cite{tan} & 54.5 & 62.5 & 65.8 & 69.0 & 70.0 \\
				Fetcher & \textbf{69.0} & \textbf{80.5} & \textbf{85.2} & \textbf{88.2} & \textbf{90.1} \\ \bottomrule
			\end{tabular}%
		}
	\end{table}

	We present a comparative analysis of our method with other state-of-the-art methods on the misspelled handwritten Chinese characters dataset. The results are shown in Table \ref{table:sota}. These methods are evaluated separately on the misspelled character set and right character set. Additionally, the misspelled character set is further divided into three subsets based on the error type: stroke-level error, radical-level error and structure disorder. When adopting RAN \cite{ran} as the baseline, our CDF-RAN achieves SOTA results. It significantly outperforms the latest SOTA method TAN \cite{tan} by a considerable margin of 18.7\% in the correct rate and 10.4\% in the decomposition accuracy of misspelled characters. Moreover, the previous methods exhibit a low precision and high recall rate on the right set, which implies that numerous misspelled samples are wrongly classified as right characters. Consequently,  the recall rate on the misspelled set remains low. Thanks to counting information, our model achieves a higher recall rate on the misspelled set and higher precision on the right set, which indicates that our method alleviates the bias problem effectively. Notably, the DACC of the right and misspelled set is both improved, which indicates that our method is beneficial to character decomposition regardless of its correctness.
	
	We further apply our approach to the latest SOTA method TAN \cite{tan}. The resulting CDF-TAN model is considerably superior to its baseline. This indicates that our approach can be generalized to various existing encoder-decoder models and enhance their performance. And it also proves that our method is useful for decomposing characters into both IDS sequences and radical-tree layouts.
	
	To demonstrate the superiority of our method in predicting ideal characters, we conduct experiments by replacing the fetcher in the CDF-RAN with existing algorithms. The results are presented in Table \ref{tab:ideal}. Given that the characters are written without context, we select the top-$k$ characters as the candidate ideal set. Edit-Distance based algorithm is employed in \cite{ran}. It calculates the edit distance between the decomposed IDS sequence and all IDS sequences of right characters, and then selects the character with the minimum distance as the ideal character. Probability-Embedding-based algorithm \cite{tan} is similar to the Edit-Distance based one. It assigns smaller weights to radicals that are deeper in the tree (See Figure \ref{img:background_decomposition} (b)) based on the assumption that as the depth of the radical increases, its influence on the whole character decreases. In contrast, our fetcher employs a completely different strategy. It predicts the ideal character directly. The fetcher outperforms existing algorithms by a significant margin, indicating its strong transfer learning ability.

	\subsection{Ablation Study}
	
	\begin{table}[]
		\caption{Ablation study of Chinese character decomposition with metric Decomposition Accuracy (in \%).}
		\label{tab:count_compose}
		\centering
		\resizebox{0.6\columnwidth}{!}{%
			\begin{tabular}{@{}l|ccc|c|c@{}}
				\toprule
				Method & Stroke & Radical & Structure & Misspelled & Right \\ \midrule
				RAN (baseline) & 43.4 & 67.2 & 60.9 & 57.3 & 94.5 \\
				+ Attention Regularization & 45.9 & 70.9 & 61.8 & 60.4 & 94.6 \\
				+ Counting Vector & 51.8 & 72.9 & 65.3 & 64.0 & 94.6 \\
				+ Re-weight Inference & 58.0 & 76.1 & 65.3 & 68.4 & 94.6 \\ \bottomrule
			\end{tabular}%
		}
	\end{table}
	
	\begin{table}[]
		\caption{Ablation study of  ideal character prediction with metric Ideal Accuracy (in \%).}
		\label{tab:ideal_ablation}
		\centering
		\resizebox{0.5\columnwidth}{!}{%
			\begin{tabular}{@{}l|ccc|c@{}}
				\toprule
				Method & Stroke & Radical & Structure & Misspelled \\ \midrule
				Attention Mechanism & 71.4 & 53.0 & 57.5 & 60.7 \\
				+ Gradient Block & 95.7 & 83.3 & 97.5 & 88.7 \\
				+ Random Drop & 96.5 & 85.3 & 97.5 & 90.1 \\ \bottomrule
			\end{tabular}%
		}
	\end{table}
	
	\textbf{Chinese Character Decomposition} To deal with the linguistic information when decomposing Chinese characters, we utilize a counter to assist the decoder. The influence of the counter can be summarized into three components: Attention Regularization, Counting Vector and Re-weight Inference. Specifically, Attention Regularization involves the inclusion of the attention regularization loss $L_{\text r}$ (See Eq. \ref{equ:L_r}) and the counting loss $L_{\text c}$ (See Eq. \ref{equ:L_c}) in the overall objective function during training. Counting Vector refers to employing the variable vector $\boldsymbol{\mathcal{C}}_t$ in the decoder (See Eqs. \ref{equ:p_yt} $\sim$ \ref{equ:c_t+1}). Re-weight Inference denotes recalculating the probability distribution during decoding in the testing stage (See Eq. \ref{eq:reweight}). To evaluate the effectiveness of each component, we plug them into the baseline model RAN \cite{ran} one by one, and evaluate the performance using the metric DACC. The results are presented in Table \ref{tab:count_compose}. Attention Regularization leads to a slight improvement in the character decomposition, since radical counting serves as an auxiliary task and contributes to the location of radicals. Counting Vector and Re-weight Inference provide counting information to the decoder and enable it to leverage linguistic information, thereby enhancing the DACC.
	
	\textbf{Ideal Character Prediction} The key to ideal character prediction lies in reducing the difference between the distributions of source and target domain data.  Therefore, we employ two strategies besides the attention mechanism in the fetcher: Gradient Blocking and Random Drop. Specifically, Gradient Blocking means blocking the gradient flow between the fetcher and other modules during training. Random Drop refers to dropping the radicals randomly when calculating the attention weights (See Eq. \ref{equ:beta}). In Table \ref{tab:ideal_ablation}, we evaluate the effectiveness of each strategy with the metric IACC.  We can observe that Gradient Blocking is necessary for achieving high performance, since it prevents the fetcher from overfitting to source domain data. Additionally, Random Drop also leads to a slight improvement by masking out different positions on the attention maps, promoting a more comprehensive understanding of the entire character.
	
	\begin{table}[]
		\caption{Ablation study of radical counting with metric Mean Absolute Error (MAE) and Mean Squared Error (MSE). We multiply all values by a hundred for better readability.}
		\label{tab:count_ablation}
		\centering
		\resizebox{0.5\columnwidth}{!}{%
			\begin{tabular}{@{}ccc|cc|cc@{}}
				\toprule
				\multirow{2}{*}{One Step} & \multirow{2}{*}{Two Step} & \multirow{2}{*}{\makecell{Attention \\ Regularization}} & \multicolumn{2}{c|}{Misspelled} & \multicolumn{2}{c}{Right} \\ \cmidrule(l){4-7} 
				&  &  & MAE & MSE & MAE & MSE \\ \midrule
				\checkmark &  &  & 0.54 & 4.68 & 0.29 & 2.62 \\
				& \checkmark &  & 0.45 & 4.40 & 0.25 & 2.49 \\
				& \checkmark & \checkmark & 0.42 & 4.15 & 0.23 & 2.28 \\ \bottomrule
			\end{tabular}%
		}
	\end{table}
	
	\textbf{Radical Counting} To achieve precise counting and accurate localization, we decouple the task of radical counting into 2 steps: (1) predicting probability vectors (See Eq. \ref{equ:p_n}) and (2) predicting counting vectors (See Eq. \ref{equ:c_n}). To verify the advantage of this decoupling, we conduct an ablation study by removing the prediction of probability vectors and directly generating counting vectors. The results are presented in Table \ref{tab:count_ablation}, which demonstrates a significant reduction in performance when counting is performed in only one step. Additionally, to find out the impact of attention regularization on the task of radical counting, we train the counter without the decoder and fetcher, and compare it with the proposed model. As shown in Table \ref{tab:count_ablation}, we can observe that the attention regularization enhances performance, indicating that the task of character decomposition can be beneficial to radical counting.

\begin{CJK}{UTF8}{gkai}	
	\subsection{Case Study}
	In this section, we select some examples to illustrate how the model accomplishes two subtasks (\textit{assessment} and \textit{correction}) of HCCEC. We compare the outcomes between RAN and the proposed CDF-RAN. The impact of linguistic information is also discussed.
	
\begin{figure*}[htb]
		\centerline{\includegraphics[width=0.7\linewidth]{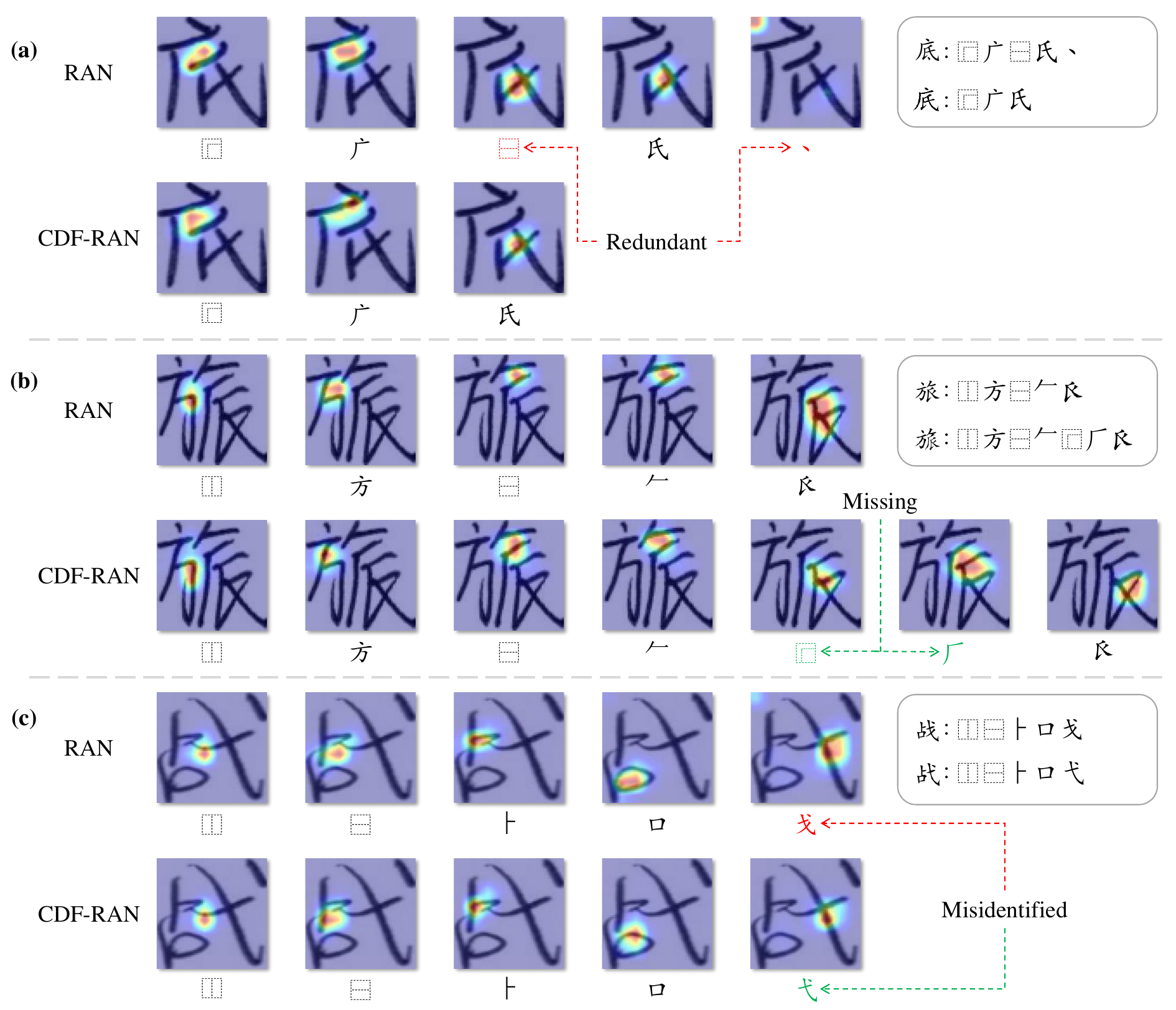}}
		\caption{Some attention visualizations from the decoder of RAN and our CDF-RAN. When processing misspelled characters, RAN mistakenly generates IDS sequences that exactly match their corresponding right ones (``底", ``旅", ``战").}
		\label{fig:attention_cv}
	\end{figure*}

	The essence of the assessment subtask lies in the decomposition of characters. We choose three images of misspelled characters and visualize their attention maps from the decoder in Figure \ref{fig:attention_cv}. Specifically, in (a), RAN erroneously generates a redundant structure ``above-below" and a radical ``丶" (marked in red) with an illogical attention map. However, it is obvious that ``丶" is absent in the image. In (b), RAN misses a structure and a radical (marked in green) that is clearly visible in the image. As for (c), although RAN locates the radical accurately at the last time step, it misidentifies the radical as a ``戈" (marked in red) instead of a ``弋" (marked in green). 
	Interestingly, these incorrect IDS sequences generated by RAN precisely match their corresponding right characters(``底", ``旅", ``战"). This is not surprising since the training data does not include these misspelled characters but only their corresponding right ones, thereby resulting in a bias towards seen IDS sequences. Thanks to the counting information from the counter, our CDF-RAN is capable of addressing the bias and accurately predicting these IDS sequences. 
	
\begin{figure*}[htb]
		\centerline{\includegraphics[width=0.7\linewidth]{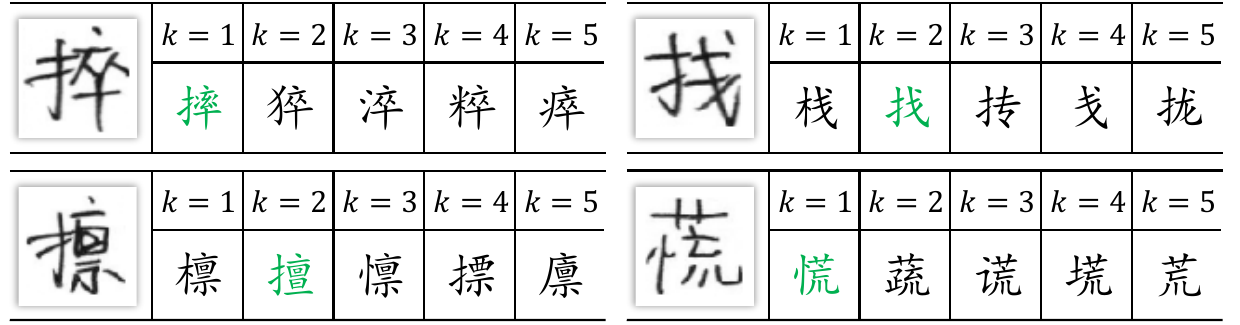}}
		\caption{Some ideal character predictions of our CDF-RAN. We present top-$k$ candidates for each case. The green character denotes the ground truth.}
		\label{fig:fetch_case}
	\end{figure*}

	Regarding the correction subtask, we present some qualitative results of the proposed CDF-RAN in Figure \ref{fig:fetch_case}. We display the five most probable candidates for each sample. The green character denotes the ground truth. Through observation, we can draw the following conclusion: (1) The ideal character is often ambiguous, which allows for multiple reasonable predictions. And it clarifies why we set top-$k$ candidates for the prediction of ideal characters. (2) Misspelled characters usually differ from their ideal characters in one or several radicals, which highlights the advantages of modeling characters at the radical level. (3) Compared with the misspelled characters, the candidates preserve radicals with strikingly glyphic information, indicating the effectiveness of the attention mechanism employed in the fetcher.
	
\end{CJK}

\begin{table}[]
	\caption{Comparison on parameters and Frames Per Second (FPS).}
	\label{table:speed}
	\centering
	\resizebox{0.27\columnwidth}{!}{%
		\begin{tabular}{@{}l|cc@{}}
			\toprule
			Method & \#Params & FPS \\ \midrule
			RAN (baseline) & 7.95M & 18.59 \\
			CDF-RAN (ours) & 10.26M & 16.30 \\ \bottomrule
		\end{tabular}%
	}
\end{table}

\subsection{Inference Speed and Extra Parameters} In order to investigate the efficiency of our method, we evaluate the speed with a single Nvidia Tesla V100 GPU. The results are presented in Table \ref{table:speed}. Compared with the baseline RAN, our CDF-RAN contains additional 2.31M parameters, which are mainly brought by the counter and fetcher. The extra time cost is marginal.

\subsection{Error Analysis}
\begin{figure*}[htb]
	\centerline{\includegraphics[width=0.6\linewidth]{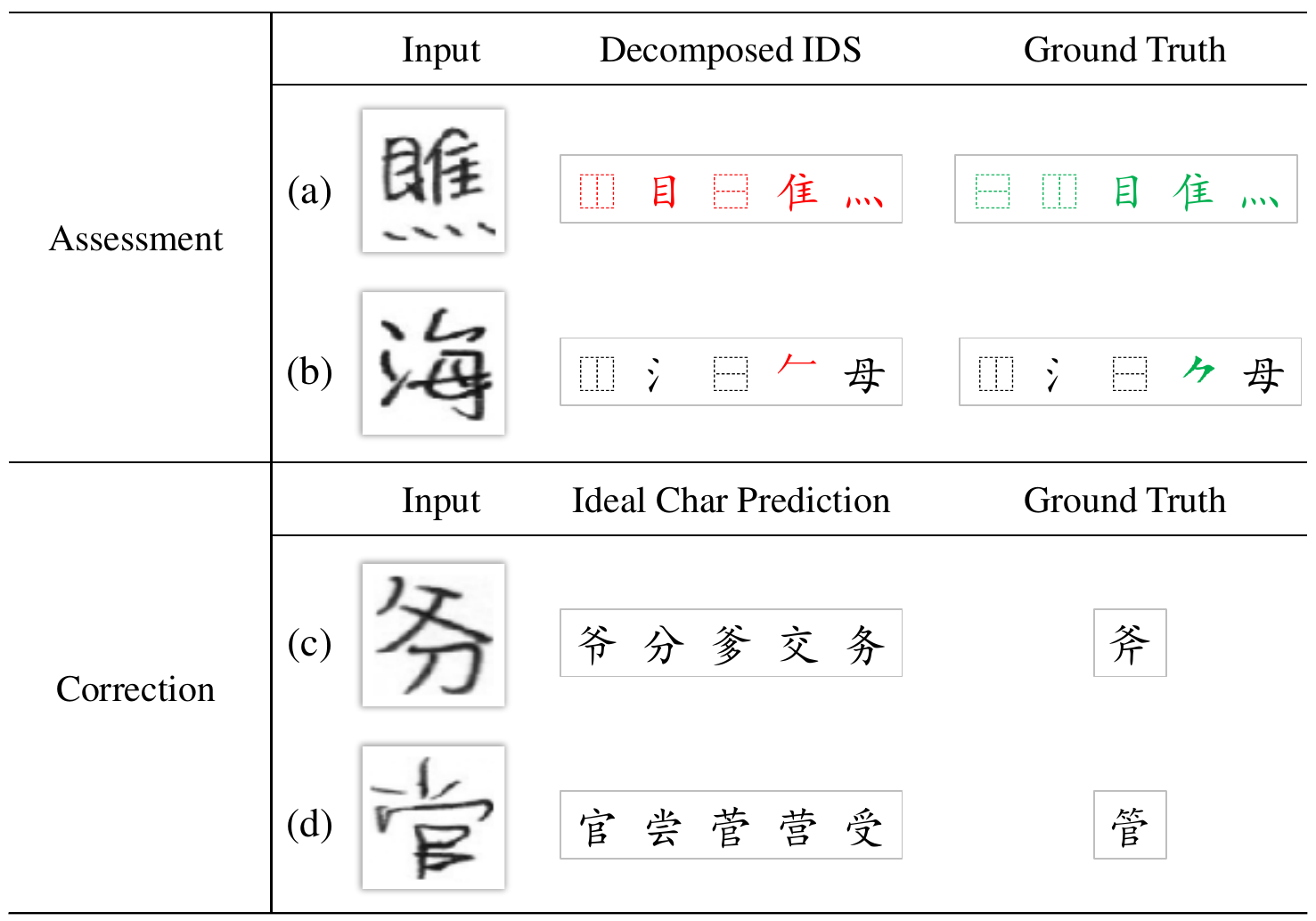}}
	\caption{Some failure cases of our CDF-RAN for the Assessment and Correction subtasks.}
	\label{fig:failure_case}
\end{figure*}
In this section, we show some failure instances encountered by the proposed CDF-RAN. As illustrated in Figure \ref{fig:failure_case}, (a) and (b) present the decomposed IDS sequence for the assessment subtask, while (c) and (d) exhibit the predictions of ideal characters for the correction subtask. Specifically, in (a), the input image presents a misspelled character. However, CDF-RAN generates an IDS sequence that matches its corresponding right character. This error arises from the fact that the misspelled character shares identical radicals and structures with its corresponding right character, thereby rendering the counting information futile. In (b), the error stems from the variations in writing styles. As for the correction subtask, the errors in (c) and (d) can be attributed to the inherent ambiguity of the task itself. In other words, these misspelled characters have multiple plausible predictions for their ideal characters, presenting a formidable challenge even for humans. Expanding the number of candidate characters could rectify such errors.

\section{Conclusion and Future Work}
In this paper, we highlight the impact of linguistic information on handwritten Chinese character error correction and present a novel approach called Count, Decode and Fetch (CDF). Through visualization and experimental results, we demonstrate that our method has the ability to deal with linguistic information properly, leading to a simultaneous improvement in the decomposition accuracy on both the misspelled character set and the correct character set. Additionally, we design a fetcher to predict the ideal character under the transductive transfer learning strategy, which overcomes the scarcity of misspelled characters for training.  In comparison to the latest state-of-the-art, our method outperforms by a remarkable margin of 18.7\% with the metric correct rate.

In future work, we will focus on two key points. Firstly, we will persist in exploring suitable methodologies to handle linguistic information, as some cases are still challenging for CDF (See Figure \ref{fig:failure_case} (a)). Secondly, we plan to enhance the visual ability of our model to distinguish similar radicals, which is not discussed in this paper.

\bibliography{reference}

\end{document}